%% file: camera_ready.tex
\documentclass[letterpaper, 10 pt, conference]{ieeeconf}  

\IEEEoverridecommandlockouts                              

\usepackage{graphics} 
\usepackage{epsfig} 
\usepackage{mathptmx} 
\usepackage{times} 
\usepackage{amsmath} 
\usepackage{amssymb}  
\usepackage{float}
\usepackage{xcolor}
\usepackage{xfrac}
\usepackage{graphicx}
\usepackage{tikz}
\usetikzlibrary{arrows, positioning}
\usepackage{leftidx}
\usepackage{hyperref}

\title{\LARGE \bf
Autonomous Hybrid Ground/Aerial Mobility in Unknown Environments
}

\author{$^*$David D. Fan$^{\dagger,\ddagger}$, $^*$Rohan Thakker$^{\dagger}$, Tara Bartlett$^{\dagger}$, Meriem Ben Miled$^{\dagger}$, \\
Leon Kim$^{\dagger}$, Evangelos Theodorou$^{\ddagger}$, Ali-akbar Agha-mohammadi$^{\dagger}$
\thanks{*These authors contributed equally to this work.}
\thanks{$^{\dagger}$NASA Jet Propulsion Laboratory, California Institute of Technology, Pasadena, CA, USA}%
\thanks{$^{\ddagger}$Institute for Robotics and Intelligent Machines, Georgia Institute of Technology, Atlanta GA, USA}%
}

\begin{document}
\maketitle
\thispagestyle{empty}
\pagestyle{empty}

\begin{abstract}
Hybrid ground and aerial vehicles can possess distinct advantages over ground-only or flight-only designs in terms of energy savings and increased mobility.  In this work we outline our unified framework for controls, planning, and autonomy of hybrid ground/air vehicles.  Our contribution is three-fold:  1) We develop a control scheme for the control of passive two-wheeled hybrid ground/aerial vehicles. 2) We present a unified planner for both rolling and flying by leveraging differential flatness mappings.  3) We conduct experiments leveraging mapping and global planning for hybrid mobility in unknown environments, showing that hybrid mobility uses up to five times less energy than flying only\footnote{Video at \href{https://youtu.be/nlGfYehTLpg}{https://youtu.be/nlGfYehTLpg}}.
\end{abstract}

\section{INTRODUCTION}
In recent years, the development of Unmanned Aerial Vehicles (UAVs) and Unmanned Ground Vehicles (UGVs) has been increasingly maturing in various applied fields, including agriculture \cite{herwitz2004imaging}, search and rescue \cite{bahnemann2017decentralized} \cite{nieuwenhuisen2017collaborative}, delivery \cite{Agha14_iros_packageDelivery}, and military applications \cite{fan2018model}.  However, the widespread adoption of autonomous vehicles still faces significant challenges, including those introduced by effective mobility in diverse environments.  
\par
While UAVs have great advantages over UGVs in terms of their ability to traverse difficult terrain, they are constrained by high energy requirements, short flight times, and low payload capabilities.  They are also fragile, non-robust to collisions, and may be adversely affected by large wind or gust disturbances.  On the other hand, UGVs can be designed to carry larger payloads, be more resistant to damage with heavier, more robust frames, and carry larger sources of power.  The trade-off is that many points of interest are often inaccessible from the ground.  Hybrid UAV/UGV vehicles are designed to combine the advantages of both classes of systems by moving on the ground to save power when possible, and flying when terrain constraints do not allow otherwise.

\par
There has been some considerable interest in hybrid ground/aerial vehicles in recent years.  The most common approach seems to be the use of passive or actuated wheels attached to the frame of a drone or quadrotor in various configurations \cite{state_1, state_2}.  Other designs include a quad-rotor hinged inside a rotating cage \cite{kalantari2014modeling}, modular or adaptive structures which change configuration for flight or ground mobility \cite{mintchev2018multi, state_8, Zhang2018}, and even legged drones \cite{li2018motion, Araki2017}.  These efforts are interesting explorations into the design space of hybrid ground/air drones.  The focus of this work, however, is a unified guidance, navigation, controls, and autonomy stack specifically geared towards Unmanned Hybrid Vehicles (UHVs).  
\par

In this work, we make use of a hybrid vehicle which consists of a quad-rotor aerial vehicle with two passive wheels attached to each side of the frame of the quad-rotor.  This design has been previously considered in \cite{kalantari2013design, kalantari2015hybrid} and has the advantage of possessing the aerial power and agility of a quad-rotor while maintaining a low additional mass profile from the added wheels.

\begin{figure}
    \centering
   \includegraphics[width=0.48\textwidth]{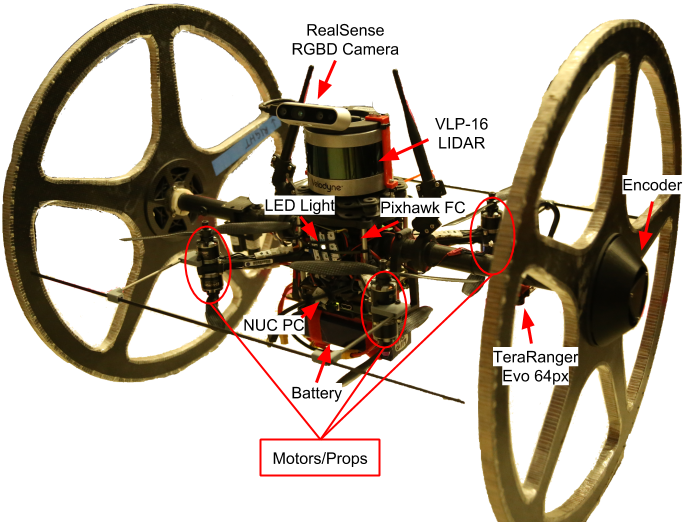}
  \caption{Rollocopter hardware configuration}
  \label{fig:rollo_sensors}
\end{figure}

The objective of this paper is to describe a unified framework for controls, planning, and autonomy of hybrid ground/air vehicles.  We take a unified planning approach which generalizes both rolling and flying by taking advantage of differential flatness mappings for quadrotors \cite{state_12} as well as for two-wheeled non-holonomic wheeled mobile robots \cite{tang2009differential}.  This results in a planner with the same structure for both mobility modes.  We present a study of autonomous navigation of our hardware platform in environments which require hybrid mobility, showing 5x energy savings of hybrid mobility vs. flying only.

\par
The main contributions of this work are as follows:
\begin{enumerate}
    \item Control of non-actuated two-wheeled hybrid quadrotor vehicles.
    \item A unified hybrid trajectory planner for both rolling and flying.
    \item Mapping and global planning for hybrid mobility in unknown environments.
\end{enumerate}

\section{System Overview}
\begin{figure}[ht]
    \centering
    \includegraphics[width=0.8\linewidth]{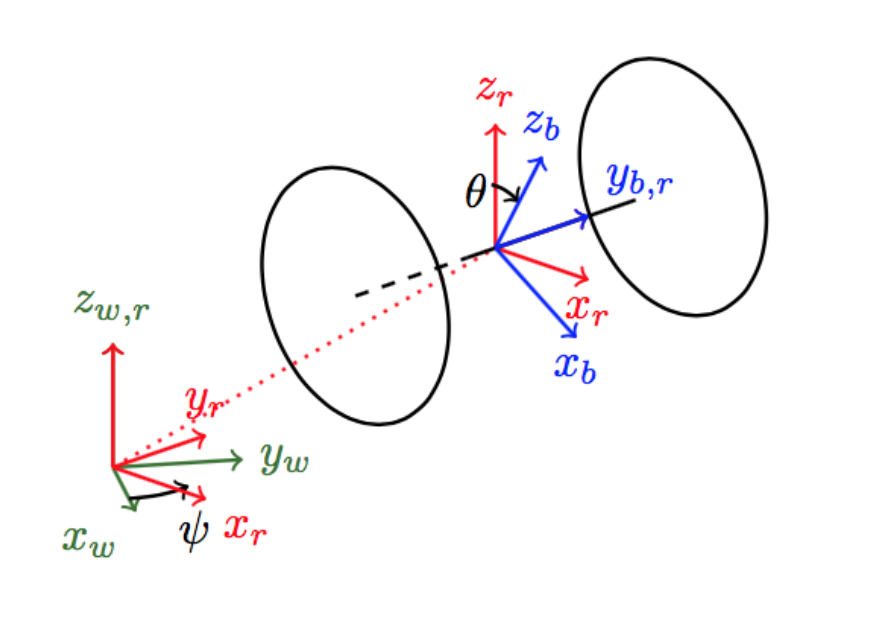}
    \caption{Figure shows reference frames: world frame $w$ (green), body frame $b$ (blue) i.e. fixed w.r.t the quadrotor frame and rolling frame $r$ (red), whose z-axis points in the direction of the surface normal of the ground and is separated from the body frame only by a rotation along y-axis.}
    \label{fig:frame}
\end{figure}

\noindent
\begin{table}
\begin{tabular}{@{\hskip0pt}p{0.2\linewidth}@{\hskip3pt}p{0.75\linewidth}}
\vspace{0.5mm}
\textbf{Notation}\\\\
$f \in \{w,r,b\}$ & Reference frames defined in Figure \ref{fig:frame}\\[0.1cm]
$s \in \{g,a\}$ &  Mobility state: $g$ ground/rolling and $a$ aerial/flying\\[0.1cm]
$^{f}\vec{p}_{g}\in \mathbb{R}^2$ & Position x, y of rolling frame w.r.t. world frame represented in $f$ frame \\[0.1cm]
$^{f}\vec{p}_{a}\in \mathbb{R}^3$ & Position x, y, z of body frame w.r.t. world frame represented in $f$ frame \\[0.1cm]
$R_{g} \in SO(2)$ & Orientation of rolling frame w.r.t. world frame \\[0.1cm]
$R_{a} \in SO(3)$ & Orientation of body frame w.r.t. world frame \\[0.1cm]
$^{f}\vec{v}_{g}\in \mathbb{R}^2$ & Linear x, y velocity of rolling frame w.r.t world frame represented in $f$ frame  \\[0.1cm]
$^{f}\vec{v}_{a}\in \mathbb{R}^3$ & Linear x, y, z velocity of body frame w.r.t world frame represented in $f$ frame \\[0.1cm]
$^{f}\vec{\omega}_{g}\in \mathbb{R}$ & Angular z velocity of rolling frame w.r.t world frame represented in $f$ frame \\[0.1cm]
$^{f}\vec{\omega}_{a}\in \mathbb{R}^3$ & Angular velocity of body frame w.r.t world frame represented in $f$ frame\\[0.1cm]
$^{f}\vec{a}_{g}\in \mathbb{R}^2$ & Linear x, y acceleration of rolling frame w.r.t world frame represented in $f$ frame  \\[0.1cm]
$^{f}\vec{a}_{a}\in \mathbb{R}^3$ & Linear x, y, z acceleration of body frame w.r.t world frame represented in $f$ frame  \\[0.1cm]
$^{f}\vec{\alpha}_{g}\in \mathbb{R}$ & Angular z acceleration of rolling frame w.r.t world frame represented in $f$ frame\\[0.1cm]
$^{f}\vec{\alpha}_{a}\in \mathbb{R}^3$ & Angular acceleration of body frame w.r.t world frame represented in $f$ frame\\[0.1cm]
$\vec{F}$ & Thrust vector whose elements represent thrust of each motor \\[0.1cm]
$\theta$ & Angle between z-axis of rolling frame and z-axis of body frame\\[0.1cm]
$\psi$ & Yaw between rolling frame and world frame \\[0.1cm]
$x^d$ & Desired state \\[0.1cm]
$\hat{x}$ & Estimated state \\[0.1cm]
$x^{ff}$ & Feedforward state
\end{tabular}
\end{table}

We first outline the overall system architecture and describe the various components.  We begin with a description of the frames used in this work.  Three frames are used which we call world, body, and rolling, the last of which is unique to the hybrid design (See Figure \ref{fig:frame}). Note that in this work we are assuming that the ground is flat, hence, $z_w$ is parallel to $z_r$.  

The full state of the quadrotor while in flight is given as $x_a = (^w\vec{p}, ^w\vec{v}, R_a, ^w\vec{\omega}_a) \in \mathbb{R}^{3} \times \mathbb{R}^{3} \times SO(3) \times  \mathbb{R}^{3}$, and the input is given by $u = (|F|,\leftidx{^b}{M}{_x},\leftidx{^b}{M}{_y},\leftidx{^b}{M}{_z})$ which maps to $\vec{F}$ i.e. the thrust of four motors.  
Similarly, the full state while rolling is given by $x_g = (^wp_g,^wv_g,R_g,^w\omega_g) \in \mathbb{R}^{2} \times \mathbb{R}^{2} \times SO(2) \times  \mathbb{R}$ while the control input is same that for flying.
\begin{figure}
\centering
\includegraphics[width=0.48\textwidth]{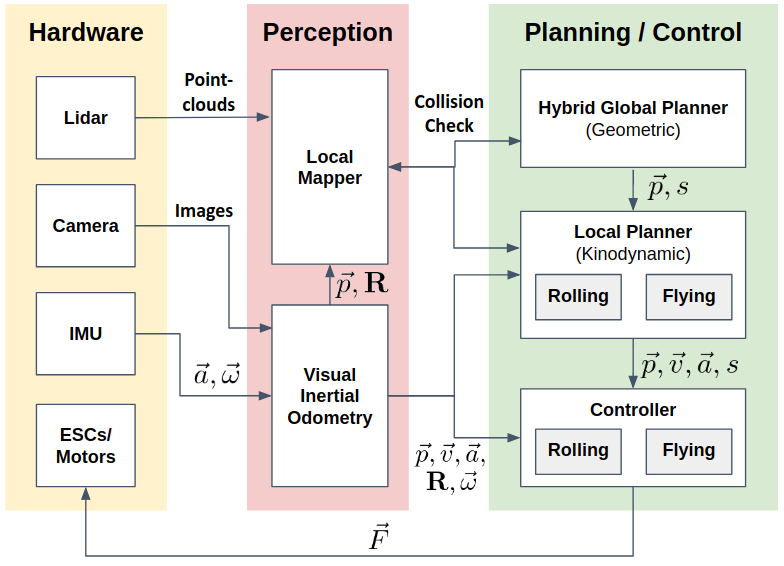}\vfill
\caption{System architecture}
\label{fig:sys_arch}
\end{figure}

As shown in Figure \ref{fig:sys_arch}, our system architecture takes a unified approach to hybrid ground and aerial autonomy.  Sensors in the hardware layer feed raw perceptual information to the perception layer.  Visual Inertial Odometry (VIO) is used to estimate the current pose.  The pose, along with pointclouds from range sensors, is used to create a map of the environment in the Local Mapper.  This map is passed along to a geometric planner, the Hybrid Global Planner, which plans hybrid trajectories in pursuit of some objective, e.g. reach some global goal or explore unexplored space.  Hybrid trajectories are generated which contain information about when the robot should roll, fly, land, takeoff, etc.  A local waypoint from this global trajectory is passed along to the lower level Local Planner.  This planner generates kinodynamically feasible trajectories which balance collision avoidance with reaching the goal.  It is able to handle both rolling and flying trajectories in a unified manner.  Once a feasible trajectory is selected, a setpoint along that trajectory is passed to the low-level Controller module.  The controller is a hybrid controller which can perform both rolling and flying maneuvers and can track given setpoints.  It sends desired attitudes/motor thrusts to the low-level flight controller or ESCs.  In the next section we describe the hybrid controller and show experimental verification of its efficacy.

\section{Controls for Flying and Rolling}
We first describe the control architecture that used for flying and then extend this architecture for ground mobility.
    
\subsection{Flying Controller}
\begin{figure}[ht]
\centering
\hspace{-1mm}\scalebox{.69}{\input{figs/control_arch2.tex}}
\caption{Architecture of the flying controller}
\label{fig:flying_controller}
\end{figure}
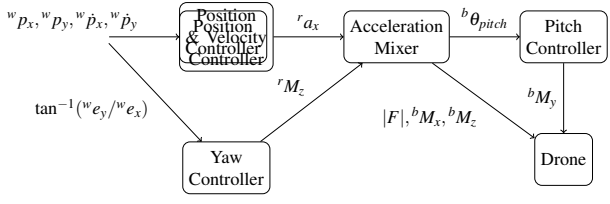

Figure \ref{fig:flying_controller} shows the non-linear back-stepping controller used for flying \cite{flying_control}, \cite{px4control}.
The controller receives the desired state $^w\vec{p},^w\vec{v},^w\vec{a},^w\psi$ from the local planner.
The feedback position controller uses a proportional control law to generate the desired velocity.
The velocity controller generates a desired acceleration based on the desired velocity from the position controller and a feed-forward term from the position controller using a PID control law.
The acceleration mixer calculates the desired thrust $|F|_1$ and attitude $\mathbf{R}$ based on desired acceleration, feed-forward acceleration and yaw.
Similar to the position controller, the attitude controller generates a desired angular velocity in body frame based on error in attitude based on proportional control law and the body rate controller uses this to generate to generate a desired moment based on a PID control law. We refer readers to \cite{px4control} for the exact equations of the dynamics and control.
\subsection{Rolling Controller}
The following are the key differences between rolling and flying modes:
\begin{itemize}
   \item In rolling mode, we assume a no-slip condition: the robot cannot have a velocity along body y-axis due to the non-holonomic constraint.
    \item The bandwidth of yaw dynamics for rolling is lower than that for flying due to interaction with the ground.
    \item The pitch dynamics for rolling are as fast as flying assuming the friction in the bearing of the wheels is negligible.
    \item The roll of the robot is constrained such that the wheels always maintain contact with the terrain while rolling.
\end{itemize}

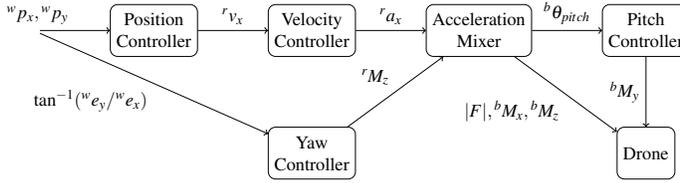
\begin{figure}[ht]
    \centering
    \hspace{-4mm}\scalebox{.69}{\input{figs/control_arch1.tex}}
    \caption{Architecture of the Rolling controller}
    \label{fig:ground_controller}
\end{figure}

Based on this intuition we extend the flying control architecture for rolling as shown in Figure \ref{fig:ground_controller}.
The following shows the equations of the control law for rolling mode.
\par
\textbf{Position Controller:}
Desired velocity along x-axis in rolling frame is calculated using a proportional control law on error in position as follows
\begin{align}
    ^we_g &= ^wp_g^d - ^w\hat{p_g} \\
    ^re_g &= R(\psi_{yaw}) ^we_g\\
    {^rv_g}_x &=  ^{pos}k_p (^re_g)
\end{align}

\textbf{Yaw Controller:}
The yaw controller generates a desired moment in rolling frame using a PD control law on $S^1$ as follows:
\begin{align}
    ^w\psi^d &= \tan^{-1}({^we_g}_y/{^we_g}_x)\\
    ^we_{\psi} &= d_{S(1)}(^w\psi^d, ^w\psi)\\
    ^rM_z &= ^{\psi}k_p  (^we_{\psi}) + ^{\psi}k_d(^w\dot{e}_{\psi})
\end{align}

\textbf{Velocity Controller:}
The velocity controller generates a desired acceleration based the desired velocity and feed-forward velocity as follows:
\begin{align}
    ^r\dot{e_g}_x &= ({^rv_g}_x^d + {^rv_g}_x^{ff} ) - {{^r\hat{v}}_g}_x\\
    ^ra_x &= ^{vel}k_d( ^r\dot{e}_x) + ^{vel}k_I\chi_1
\end{align}


\textbf{Acceleration Mixer:}
\begin{align}
    ^b\theta_{pitch}^d &= \sin^{-1}(^ra_x m / |F|)\\
    ^bM_x,^bM_z &= R(^b\hat{\theta}_{pitch})^rM_z
\end{align}

\textbf{Pitch Controller:}
\begin{align}
^be_{pitch} &= d_{S(1)}(^b\theta_{pitch}^d,^b\hat{\theta}_{pitch})\\
^bM_y &= ^{pitch}k_p(^be_{pitch})+^{pitch}k_d(-^b\dot{\theta}_{pitch}) + ^{pitch}k_I\chi_2
\end{align}

where $^{pos}k_p$, $^{\psi}k_p $, $^{\psi}k_d$, $^{vel}k_d$, $^{\theta}k_p$, $^{\theta}k_d$ and $^{\theta}k_I$ are constant gains, ($\chi_1$, $\chi_2$) the integral position tracking error of velocity and pitch angle respectively.

\begin{figure}[ht]
    \centering
       \includegraphics[width=0.48\textwidth]{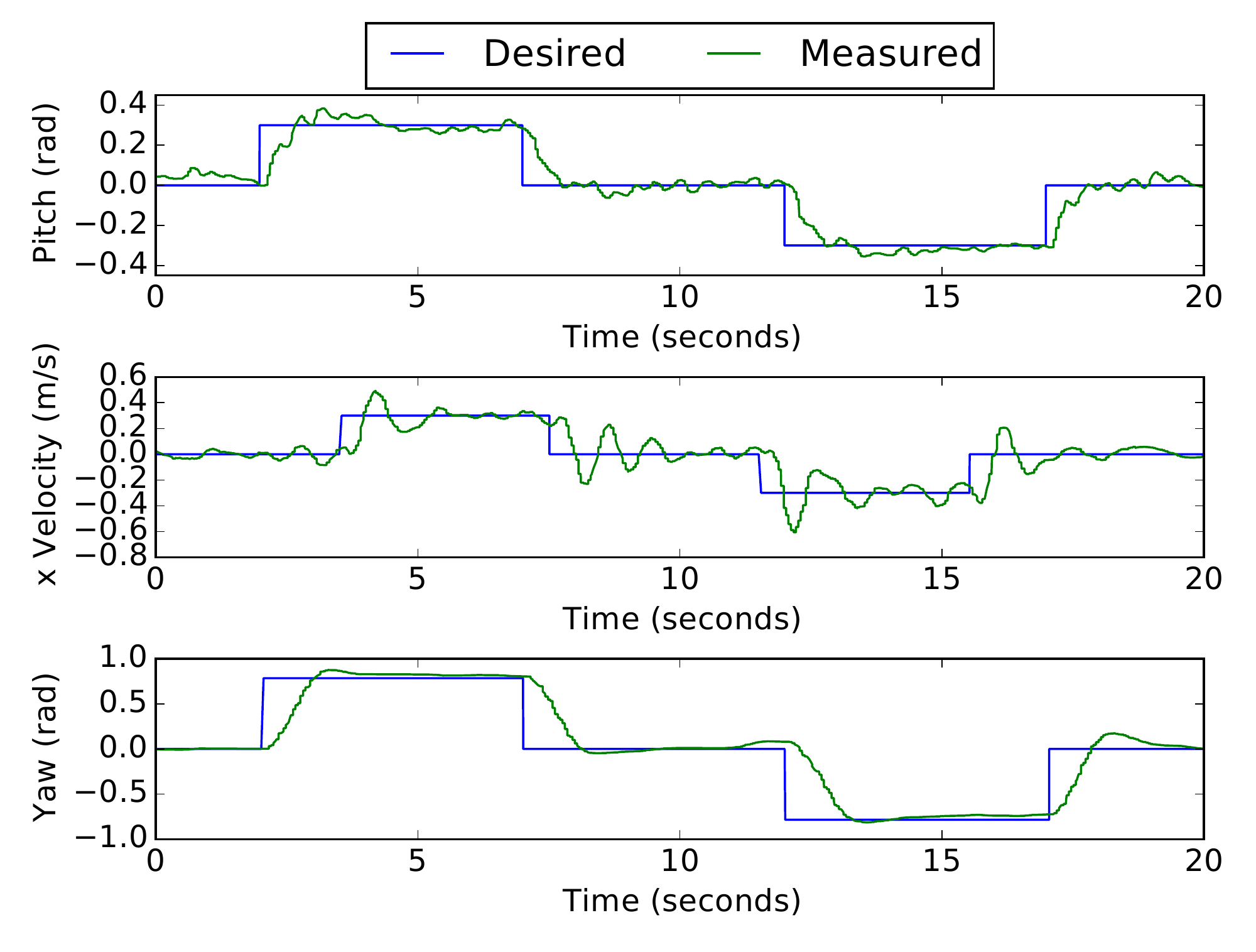}
      \caption{Step responses for the pitch, velocity and yaw controllers for rolling.}
      \label{fig:pitch_vel_yaw}
\end{figure}

\subsubsection{Experimental Validation}
We first show performance of the pitch controller while keeping the wheels fixed (Figure \ref{fig:pitch_vel_yaw}).  After tuning the pitch controller, we give desired step inputs in x velocity and plot the response.  Finally we give step inputs in yaw and tune the yaw controller.

\begin{figure}
\centering
   \includegraphics[width=0.48\textwidth]{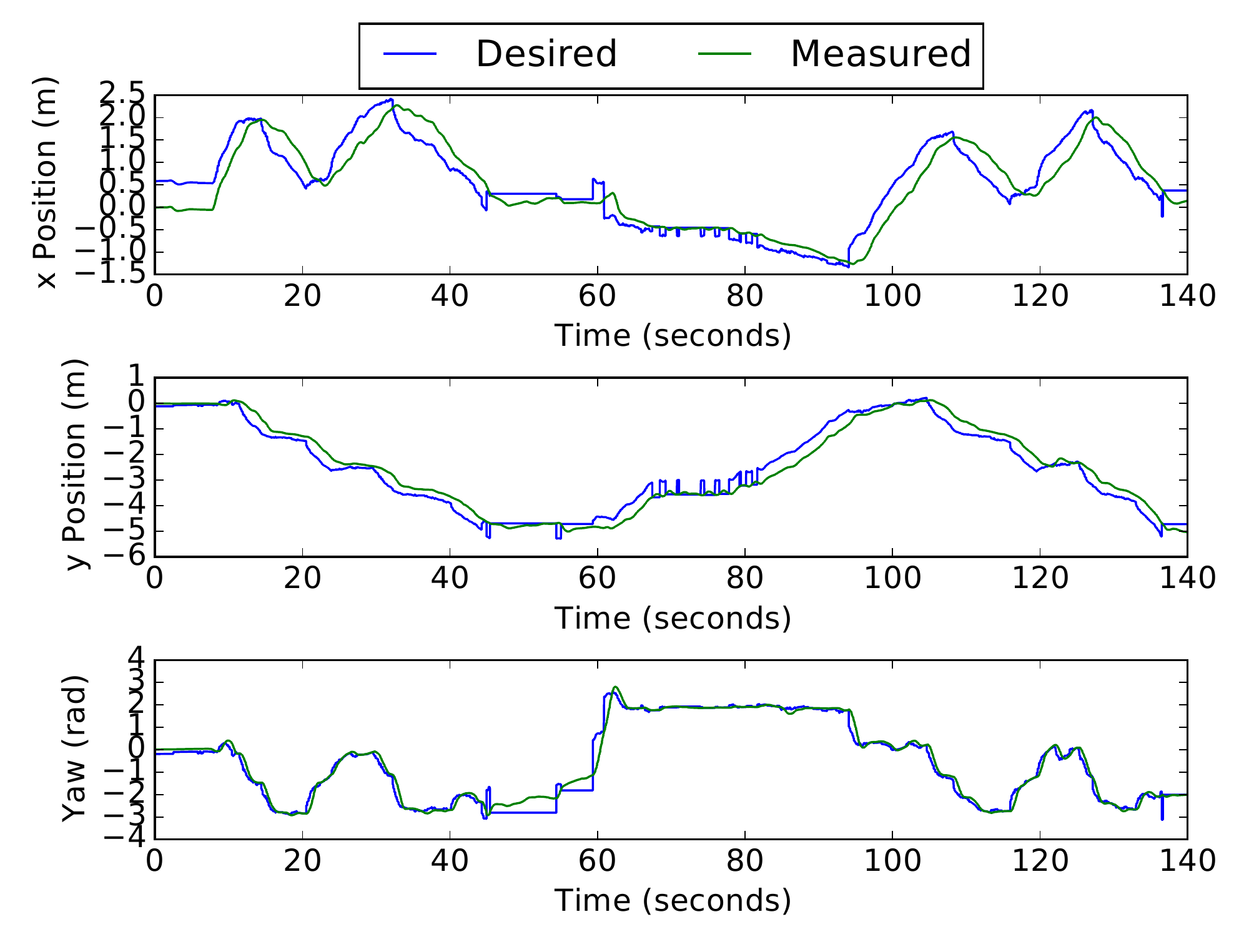}
  \caption{Performance of rolling controller in response to changing set-points given by the local planner while traversing through the test course.  Plots show position responses in x and y as well as yaw.}
  \label{fig:x_y}
\end{figure}

After tuning the controller with step responses we test the performance of rolling in a test course, as discussed in Section \ref{sec:test_course}.  This demonstrates that the robot is able to successfully navigate with course without colliding, proving the efficacy of the rolling controller.

\subsection{Transition between Rolling and Flying}

As shown in Fig \ref{fig:sys_arch} transition between flying and rolling is made by the Local planner. Besides position, velocity and acceleration, the hybrid controller receives the mobility mode $s$ : ground or aerial.

\textbf{Take-Off}:
Take-off is triggered by increasing the set-point along the z-axis in world frame and finished when the ground clearance is above a certain threshold.

\textbf{Landing}:
Landing is triggered by asking for a negative velocity along the z-axis in world frame and finished when the ground clearance is lower than a certain threshold for 1 second.

\section{Local Planning}
\subsection{Hybrid Differential Flatness Planning}
The design requirements of our local planner are as follows:  
\begin{enumerate}
    \item Accept position waypoint goals in the world frame.
    \item Plan kino-dynamically feasible trajectories which are collision-free, for both rolling and flying.
    \item Weigh competing costs of reaching the goal vs. maintaining a safe distance from obstacles.
    \item Replan at the rate of received instantaneous pointclouds for agile responsiveness to dynamic obstacles.
    \item Return a position and velocity setpoint for the controller to track.
\end{enumerate}

Various trajectory planning methods exist in both the literature and in practice.  Our approach seeks to fulfill these design requirements in a way that takes into account hybrid mobility.  To this end we use a motion primitive-based approach.  Generally motion primitives are computed via alternative paths \cite{zhang2018pcap} or by generating trajectories to a selection of waypoints within the field of view \cite{paranjape2015motion}.  In this work we take the latter approach, generating kino-dynamically feasible trajectories using the same representation for both aerial and ground mobility.

\begin{figure}[ht]
\centering
  \includegraphics[width=0.9\linewidth]{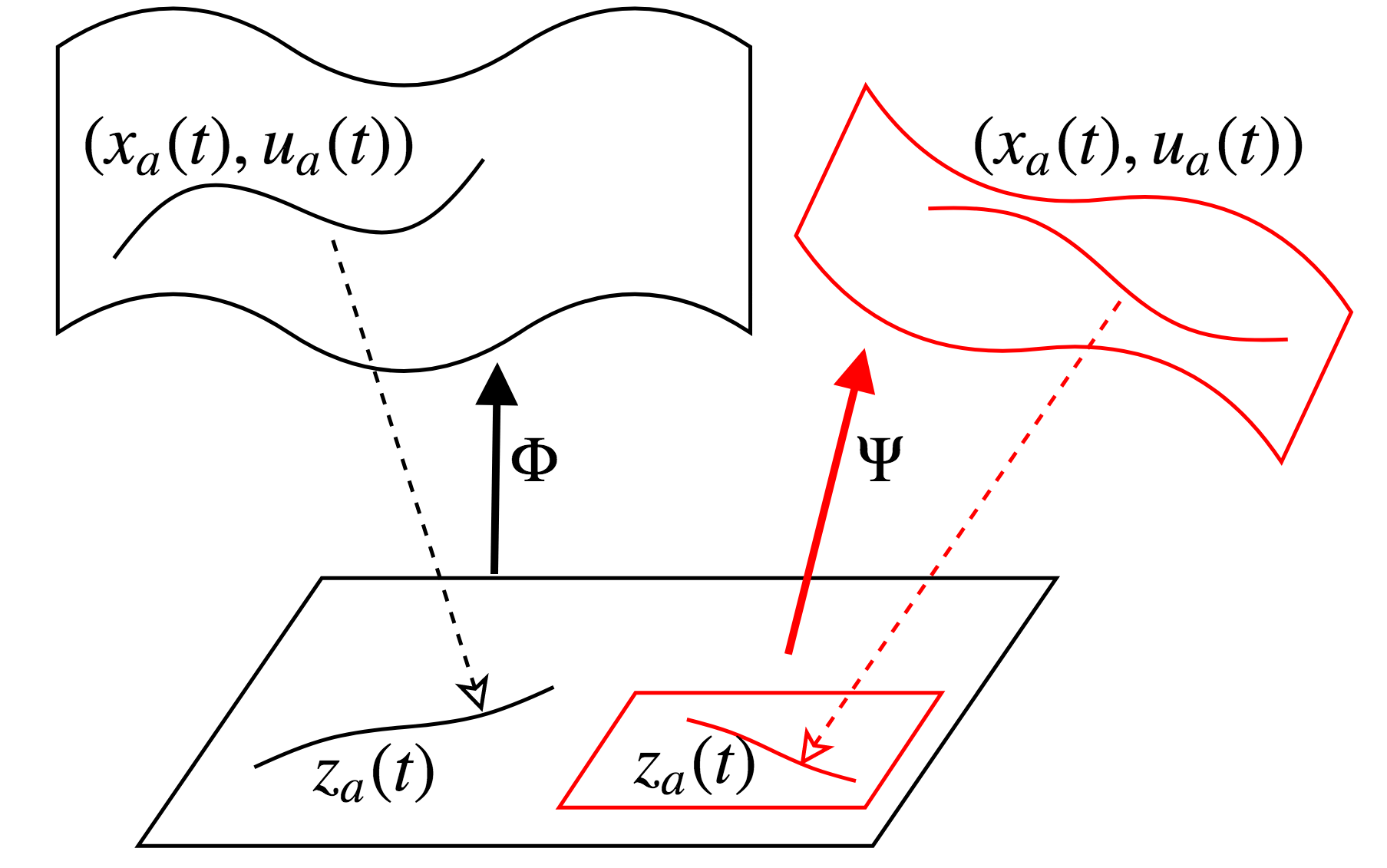}
  \caption{Differential flatness}
  \label{fig:flat}
\end{figure}

Prior work has shown that differential flatness mapping-based planning for quadrotor flight has advantages in terms of agility, computational efficiency, and simplicity of design \cite{mahony2012multirotor, lopez2017aggressive}.  Similar differential-flatness based mappings exist for a variety of non-holonomic vehicle configurations, including two-wheeled mobile robots \cite{murray1995differential, tang2009differential}.  We first begin with a brief overview of differential flatness mappings for planning, then show how the dynamics of two-wheeled mobile robot maps to a subspace of the flat space of quadrotor dynamics (Figure \ref{fig:flat}).  This enables us to plan using the same representation for rolling and flying, leading to a unified hybrid local planner.

It has been shown \cite{mahony2012multirotor} that one can find a mapping from the state $x_a$ and control $u$ to a flat space consisting of four variables: $z_a=(^w\vec{p}_a,\psi)$ and the derivatives of $z$.  Define a trajectory $z_a(t):[t_0,T]\rightarrow \mathbb{R}^3 \times SO(2)$ as a smooth curve parameterized by $t$.  There exists  a smooth mapping $\Phi$, such that $(x_a,u)=\Phi(z_a,\dot{z}_a,\ddot{z}_a,\dddot{z}_a,\ddddot{z}_a)$. 
This is the differential flatness mapping from the dynamics of the quadrotor in flight to a flat space.  Prior work has shown that one can construct minimum-snap polynomial trajectories in this flat space which map back to trajectories of the quadrotor dynamics \cite{mahony2012multirotor}.  Here we take a similar approach and define polynomial trajectories by their start and end points while assuming the accelerations, jerks, and snaps at these points are 0, i.e. $z_a(t)=(\vec{p_a}(t),\psi(t))$, $\dot{z}_a(t)=(^w\vec{v},^w\dot{\psi(t)})$, $\ddot{z}_f(t)=\dddot{z}_f(t)=\ddddot{z}_f(t)=0$ for $t=t_0$ and $t=T$.

We now review a similar deferentially flat mapping for two-wheeled mobile robots \cite{tang2009differential}. 
When the vehicle is rolling on flat ground, the state representation is reduced to $x_g=(\vec{x}_g,\psi)$ and the control inputs are $u_g=({^rv_g}_x,\dot{\psi})$.  
There exists a differential flatness mapping from $x_g$ and $u_g$ to the flat space $z_{g}=(^w\vec{p})$, i.e. $(x_g,u_g)=\Psi(z_g,\dot{z}_g)$.  
Note that $z_g$ is a subspace of $z_a$.  We therefore can plan trajectories for rolling using the same polynomial representations we use for generating flight trajectories.  
Instead of generating 4-dimensional polynomials in $z_a$, we generate two-dimensional trajectories in $z_r$. 
In this way we maintain a consistent architecture and pipeline for both mobility modes.

\begin{figure}[ht]
\centering
  \includegraphics[width=0.9\linewidth]{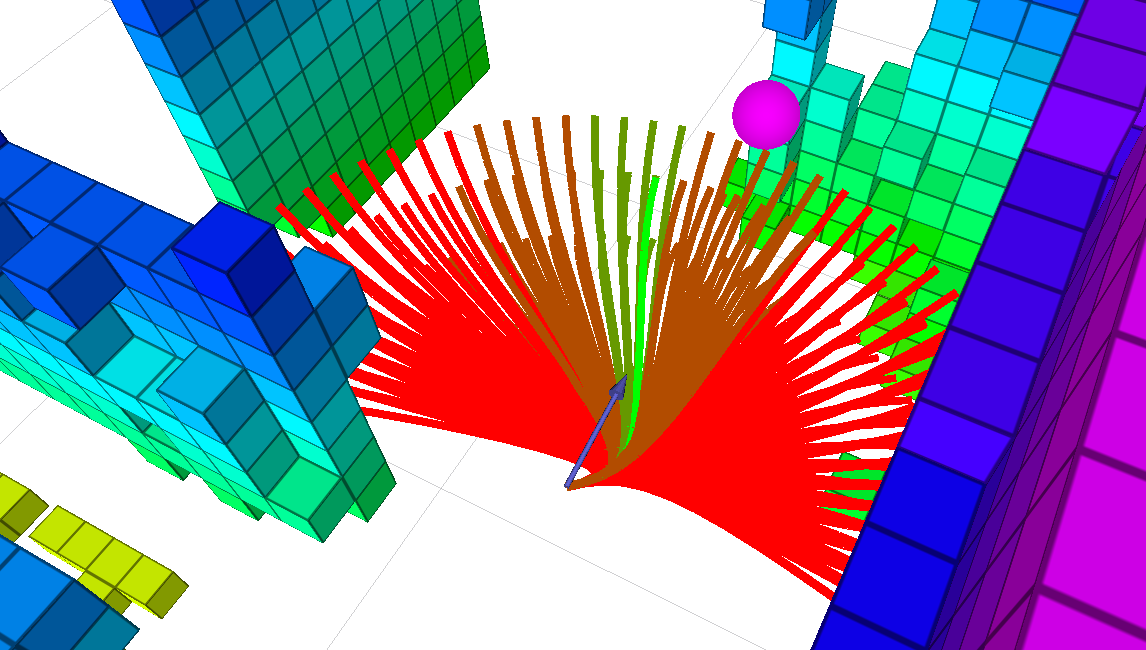}
  \caption{Motion primitives radiating outward from the origin of the vehicle, with color-coded costs.  Red represents primitives which are in collision.  Dark red are primitives within the near-collision buffer.  Dark green primitives are free-space primitives.  Bright green indicates the lowest cost primitive.  Note that primitives radiate outwards in x,y and z.  The pink sphere is the local goal, and the blue arrow is the current pose of the vehicle.  Colored boxes represent occupied space with color representing the height.}
  \label{fig:motion_primitives}
\end{figure}

For both mobility modes, primitives are created based on a selection of endpoints. An arc of endpoints are created in the horizontal plane around the vehicle are determined based on a given distance horizon. In the aerial mode, additional endpoints at different heights are included at a variety of angles above and below the horizontal plane as seen in Figure \ref{fig:motion_primitives}.

\subsection{Selecting Optimal Trajectories}
After generating a set of possible kinodynamically feasible trajectories, the planner needs to select the best one.  The motion primitives are checked for collisions by using multiple query points along each primitive.  For each query point, the distance to the nearest collision point in the representation of the environment is determined using a KD-tree search.  
If the distance to the nearest obstacle $d_{obstacle}$ is less than the  radius of the vehicle, the trajectory is considered to be in collision and rejected by assigning it a large collision cost $C_{collision}=10^6$.  On the other hand, if $d_{obstacle}$ is larger than some buffer value, say 1 meter, then we consider it to be in \textit{near collision} and it is given a collision cost of $C_{collision}=100 \times W_g - d_{obstacle}$, where $W_g$ is a constant factor which weighs moving towards the goal against staying away from obstacles. We then assign the total cost of the primitive as 
\begin{align*}
C_{goal} &= W_g \times d_{goal}\\
C_{total} &= C_{collision} + C_{goal}
\end{align*}
\subsection{Choosing a controller setpoint}
The lowest cost primitive is used to generate a setpoint for the controller.  This setpoint can be the point along the trajectory corresponding to the current time.  When replanning with a high frequency, this setpoint can be a fixed time interval ahead of the initial time at which the trajectory begins.  

When the local planner receives a new goal, that goal has a mobility mode associated with it.  If the mobility mode of the goal is different from the current mobility mode, the local planner will switch mobility modes, either from rolling to flying or vice versa.  The local planner will then send a flying controller setpoint to command takeoff or landing.  The takeoff command is achieved by sending an increasing z position setpoint to the controller. This transition is declared complete when the ground clearance is sufficiently high. Similarly, the landing command is sent to the controller as a negative velocity in the z direction. When the ground clearance is approximately equal to the wheel radius for a specified time interval, the vehicle is declared as having landed.

\subsection{Localization-free robustness}
In GPS-denied environments, or where position information is not available (i.e. $\vec{p}$ is unknown), position control is no longer feasible.  A common approach is to perform collision avoidance on instantaneous collision information and follow walls (see \cite{katsev2011mapping, van1992wall, santos1995divergent, serres2006toward, roubieu2012fully}, and many others).  Our hybrid motion primitive approach is well-suited to this type of behavior.  When localization is unavailable, we send desired velocities only, since velocity estimates are more reliable even when position-based localization fails.  Because the differentially flat trajectories are very cheap and easy to generate, it is possible to re-plan at a high frequency, even at the frequency of instantaneous collision point clouds.  Even without mapping and localization, basic collision-avoidance behaviors emerge from this framework.  This provides a level of robustness to unreliable localization methods.  More discussion about localization and mapping follows in the next section.

\section{Localization, Mapping, and Global Planning}
    \subsection{Localization}
    The ability to safely navigate and traverse unknown and GPS-denied environments necessitates a reliable autonomous localization system that can run in real-time. The system utilizes ORB-SLAM2 with RGB-D data from the Intel Realsense cameras for pose estimation. ORB-SLAM2 is an open-source SLAM system for monocular, stereo, and RGBD cameras which leverages the speed of ORB features and back-end bundle adjustment to enable lightweight and accurate real-time localization \cite{Mur-Artal2017}. This pose estimate $(\vec{p},\psi)$ is fused with the IMU onboard the Pixhawk flight controller via an onboard EKF to produce state estimates.

    \subsection{Mapping}\label{sec:local_mapping}
    Mapping for mobile robots is a well-studied problem \cite{state_25}.  In this work we take a traditional approach and leverage existing work. The point clouds from a 3D LIDAR along with the state estimates are combined into a local map using OctoMap that can be used for higher fidelity planning and navigation in geometrically complex environments.  OctoMap is an open-source mapping framework which utilizes the octree data structure and probabilistic occupancy estimation to efficiently generate three-dimensional models of the environment \cite{Hornung2013}.  A local map of obstacles is generated in a 5m $\times$ 5m $\times$ 3m region around the current position, and is cleared upon localization failure. 
     The hybrid mobility of the vehicle makes mapping especially important due to the limited field of view in elevation angle of the VLP-16 LIDAR as well as occlusion by the wheels. 
     
    \subsection{Hybrid Global Planning}
\begin{figure*}[ht]
    \centering
       \includegraphics[width=\textwidth]{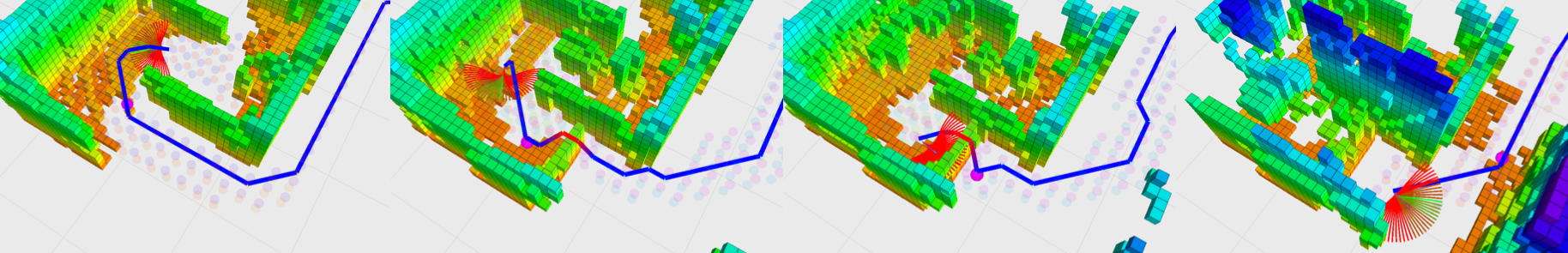}
      \caption{Time sequence of planned hybrid transition from rolling to flying and back to rolling.  Colored blocks are occupied voxels colored by z height.  Blue/Red path indicates hybrid planned path where blue is rolling and red is flying.  Transparent spheres denote A* nodes.  Pink sphere is goal waypoint sent from the global planner to the local planner.  Motion primitives from local planner are shown.  Note that as the vehicle moves forward, an obstacle is revealed and a small hop over it is planned and executed.}
      \label{fig:planning}
    \end{figure*}

    The vehicle autonomously navigates through the freshly mapped environment to a goal position through the use of an A* search planner \cite{Edwards1963}.  Nodes of the search graph are placed in free space.  
     An incrementally updated Euclidean Distance Transform (EDT) is calculated from the occupancy map through a brushfire algorithm to determine whether a node is in collision or close enough to an obstacle to be considered collided  \cite{Lau2013}.  Each node has a cost which is determined by its distance from obstacles.  An additional cost for flight is added to nodes which are above the ground.  This cost is formulated to account for the increased energy required to fly compared to rolling and results in paths which prioritize rolling over flying when the terrain allows for it.  See Figure (\ref{fig:planning}).
     \subsection{Local Waypoint Generation}
     Given the hybrid path planned by A*, the global planner sends a goal to the local planner.  This goal is chosen by taking the $n^{th}$ node further along the path from the current position.  If the next node in the path is a different mobility mode than the current state of the vehicle, then a transition is requested of the local planner and controller, either from rolling to flying or vice versa.  The replanning frequency is determined by octomap updates and the length of time taken to perform A* search, generally on the order of 1-2Hz.
     
\section{Experiments}
In this section, we show experiments to compare energy consumption for different mobility modes of the rollocopter.

\subsection{Hardware}
Figure \ref{fig:rollo_sensors} shows the hardware configuration of the Rollocopter system.  The electronic speed controllers (ESCs), motors, and propellers are used to control the drone. The wheels, connected to the drone by the main shaft, are made of a light carbon-fiber honeycomb structure. The vehicle is actuated by the propellers and uses the wheels for rolling, thus enabling hybrid mobility.  The platform uses the following sensors onboard:
\begin{itemize}
    \item RealSense RGB-D Camera: This camera provides the stereo and RGB-D images which can be used for simultaneous localization and mapping algorithms, such as ORBSLAM.
    \item Velodyne 360$^o$ VLP-16 LIDAR: The LIDAR provides a point cloud around the vehicle with a 360$^o$ azimuth angle and $\pm15^o$ elevation angle field of view. This is used for collision checking and local mapping.
    \item IR LED-based height sensor (TeraRanger Evo 64px):  Two TeraRanger Evo 64px height sensors \cite{TeraRanger} are used to provide bottom and top clearances.  These sensors are critical for mitigating the limited field of view of the Velodyne sensor in the downwards and upwards directions.
    \item Pixhawk flight controller: We use a Pixhawk flight controller with onboard IMU, gyroscope and attitude estimation.
    \item Encoders: We utilize wheel encoders to aid in velocity estimation while operating on the ground.
    \item Intel NUC: The system is equipped with onboard computing, an Intel i7 Core processor and 32GB RAM.  The system runs Ubuntu 16.04 and we use ROS \cite{ros} for message passing and handling. 
    
\end{itemize}
The total mass of the system is 4.231 kg.  The wheels, bearings, and mounting, weigh around 500g.  We experimentally verify a maximum thrust rating of 62.5 N.

\subsection{Test Environment}\label{sec:test_course}
\begin{figure}[ht]
\centering
   \includegraphics[width=0.35\textwidth]{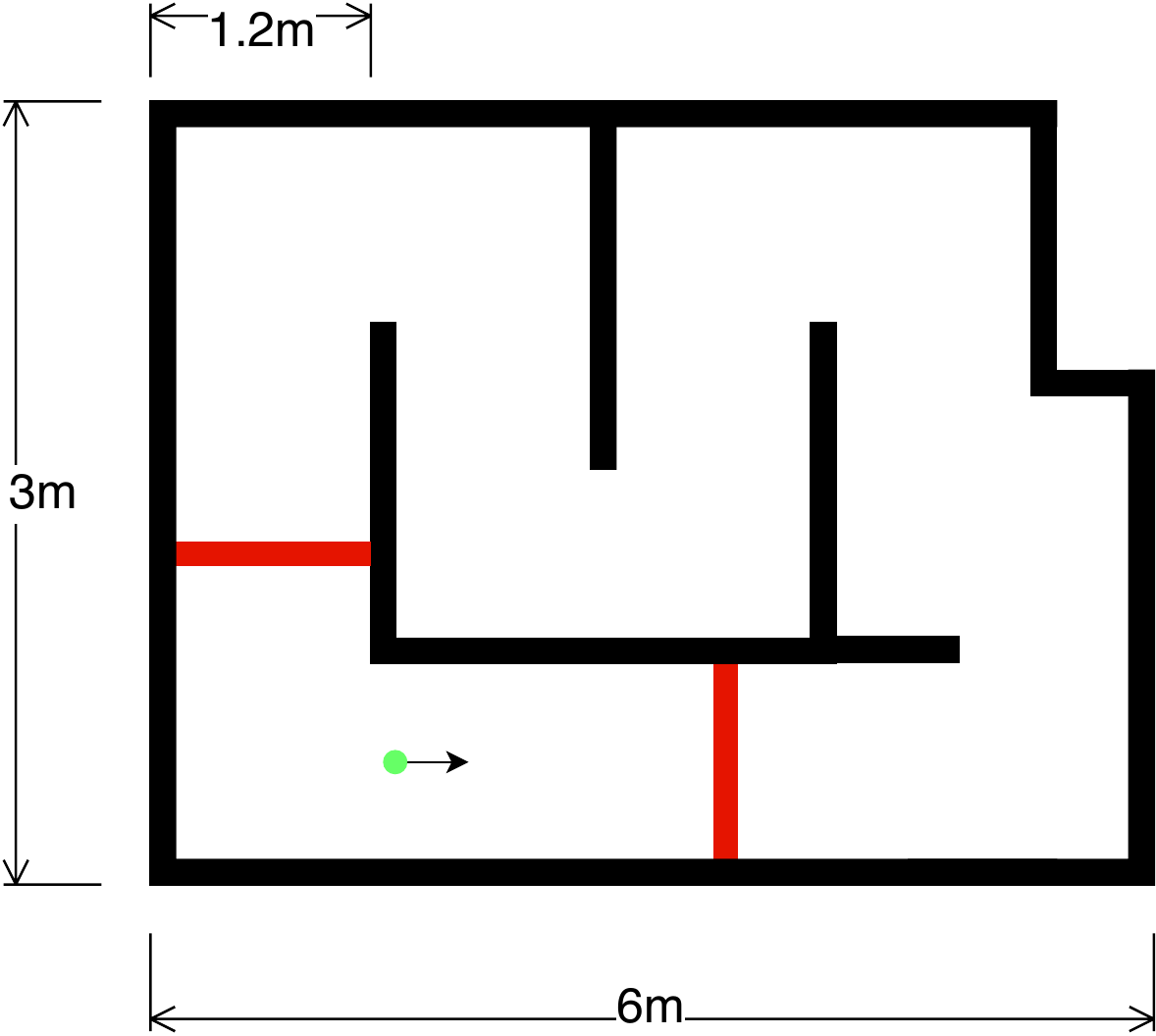}
  \caption{Top-down view of test course with walls represented by black lines. The red lines represent removable obstacles. The green point and arrow represent the starting pose of the vehicle.}
  \label{fig:test_course}
\end{figure}

The test course in Figure \ref{fig:test_course}is suitable for both flying and rolling and contains narrow corridors and hairpin turns.
  It also contains removable obstacles for testing hybrid mobility.  The walls are 1.5m high and removable obstacles are 0.4m high.  The top of the course is covered to create an enclosed maze.  The total length of one circuit around the track is approximately 18m.  Due to the narrow width of the corridors (1.2m) and the frequent 180$^o$ turns, a high degree of collision avoidance, and mobility is required.

\subsection{Energy Comparison of Different Mobility Modes}
Our first experiment was a comparison between rolling and flying.  We collected rolling and flight data in the test course and log battery voltage and current levels.  The desired velocity of both air and ground traversal was set at 0.3 m/s.  A comparison between the average power used during both mobility modes is found in Table \ref{table:power}.  In these experiments we found that on average, rolling is ~5x more efficient than flying.  This is a conservative lower bound, and further gains in energy efficiency while rolling should be possible, since we did not dynamically adjust the thrust of the rolling controller.  We leave this for future work.

As an initial estimate of the hybrid vehicle's energy efficiency over a purely aerial vehicle of a similar configuration/mass without the hardware needed to roll, the average power consumption of both cases can be compared. While we experimentally verified the power consumption of the vehicle with wheels in Table \ref{table:power}, we approximate the average power consumption of an aerial vehicle without wheels to be roughly 650 W. From this approximation and our experimentally determined average power consumption values, the minimum percentage of rolling time to total travel time to justify a hybrid vehicle configuration over a purely aerial one is calculated as 41.4\%. This value is obtained under the assumption that the average power consumption of the hybrid vehicle is a time-weighted average of the experimentally determined power consumption values of rolling and flying. 

\begin{table}[ht]
  \centering
\begin{tabular}{ | c | c | c |} 
\hline
Mobility Mode & Avg. Power (Watts) & $\sigma$\\
\hline
Rolling & 194.5 & 17.6 \\
Flying & 971.9 & 66.3 \\
\hline
\end{tabular}
\caption{Comparison of average power usage during rolling vs. flight over 7 minutes.  Rolling is ~5x more efficient.}
\label{table:power}
\end{table}

\begin{figure}[ht]
   \includegraphics[width=0.48\textwidth]
   {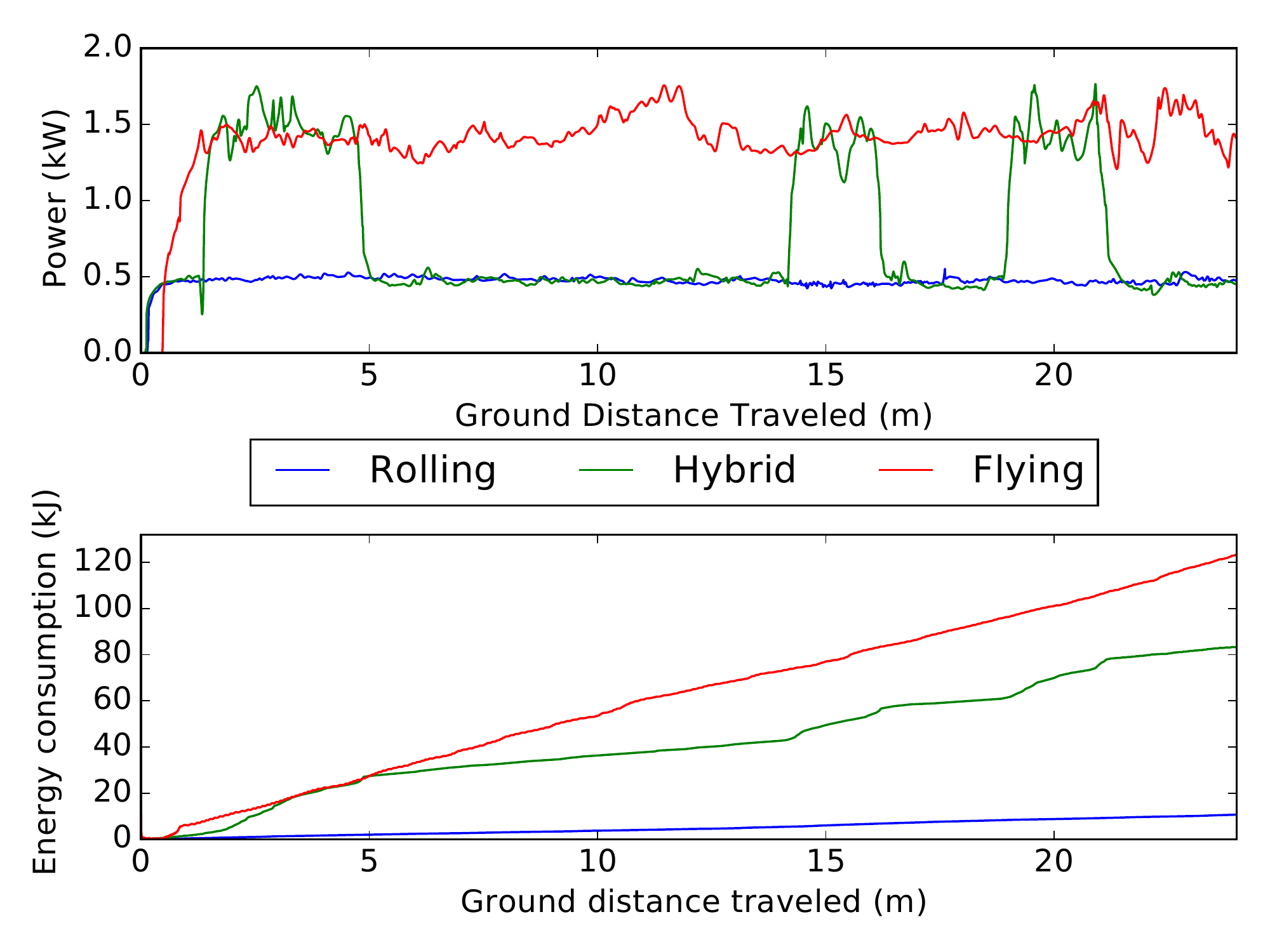}
  \caption{Comparison of power and distance capabilities of the hybrid platform while rolling, hybrid planning, and flying in the test course.  The hybrid planner flies over three obstacles, while rolling was tested with all obstacles removed.}
  \label{fig:energy}
\end{figure}

Next, we conducted experiments to quantify the power consumption of the vehicle in rolling only, flying only and hybrid modes while planning autonomously in the test course. 
We first removed any obstacles to allow the vehicle to traverse the entire course without flying.  We verified that the planner chooses to roll at all times.  We plotted the power used as well as energy consumption as a function of the ground distance the vehicle travels (Figure \ref{fig:energy}.  We then add the obstacles to the course.  The hybrid planner autonomously plans trajectories through the course, flying when it encounters an obstacle, then landing again, in an energy-optimal manner (Figure \ref{fig:planning}).  Finally, we force the local planner to fly only, traversing through the course on flight alone to perform a comparison between hybrid mobility vs. flight.  

In these experiment shown in Figure \ref{fig:energy} we observed about a 3.5x reduction in power consumption while rolling vs. flying, due to the use of a higher desired thrust while rolling.  However, $>$5x power reductions are entirely feasible by using lower rolling thrust.  This reduction depends on a number of factors including wheel friction, terrain traversability, vehicle configuration, etc.  See \cite{kalantari2013design,kalantari2015hybrid,sabet2019rollocopter} for a more thorough treatment of the energy trade-offs associated with this design.

\section{CONCLUSIONS}
With the hybrid vehicle system design, we are able to show advantages in reduced energy consumption, increased maximum run times, and increased distance traveled, which translates to further exploration of unknown environments.  We have presented a novel controller for rolling, a unified planning framework for both ground and aerial mobility, and experiments quantifying the energy savings of hybrid mobility while exploring unknown, constrained environments.  Future work includes a more comprehensive analysis and design of controllers for ground mobility on rough terrain, consideration of differential flatness trajectory optimization in the context of Model Predictive Control, and investigation of other hybrid vehicle designs.

\section*{ACKNOWLEDGMENT}
The authors would like to thank the JPL SubT team for their support in the development of the hybrid aerial/ground vehicle. We would particularly like to thank the hardware team consisting of: Scott Harper, Matt Anderson, Arash Kalantari, Fernando Chavez, Kyle Strickland, and Thomas Touma. We would also like to thank Brett Lopez for discussions regarding formulating the hybrid controls and planning framework.

This research was partially carried out by the Jet Propulsion Laboratory, California Institute of Technology, and was sponsored by Ali-akbar Agha-mohammadi.  Copyright \textcopyright 2019. All rights reserved.
\bibliographystyle{ieeetr}
\bibliography{references}

\end{document}

%% file: figs/control_arch2.tex
\tikzstyle{block} = [rectangle, rounded corners, minimum width=1cm, minimum height=1cm,text centered, draw=black]

\begin{tikzpicture}[node distance = 1.35cm, auto]
\node (pseudo)[]{};
\node (pc) [block, right =of pseudo, align=center] {Position \\Controller};
\node (vc) [block, right =of pseudo, align=center] {Position \\ \& Velocity \\Controller};
\node (ac) [block, right =of vc, align=center] {Acceleration \\ Mixer};
\node (yc) [block, below =of vc, align=center] {Yaw \\ Controller};
\node (ptc) [block, right =of ac, align=center] {Pitch \\ Controller};
\node (drone) [block, below =of ptc, align=center] {Drone};

\draw [->] (pseudo) -- node[anchor=south east]
{$\leftidx{^w}{p}{_x}, \leftidx{^w}{p}{_y}, \leftidx{^w}{\dot{p}}{_x}, \leftidx{^w}{\dot{p}}{_y}$} (pc);
\draw [->] (pseudo) -- node[anchor=north east]{$\tan^{-1}(^we_y/^we_x)$} (yc);
\draw [->] (yc) -- node[anchor=south east] {$\leftidx{^r}{M}{_z}$} (ac);
\draw [->] (vc) -- node[anchor=south] {$\leftidx{^r}{a}{_x}$} (ac);
\draw [->] (ac) -- node[anchor=south] {$\leftidx{^b}{\theta}{_{pitch}}$} (ptc);
\draw [->] (ptc) -- node[anchor=east] {$\leftidx{^b}{M}{_y}$} (drone);
\draw [->] (ac) -- node[anchor=north east] {$|F|, \leftidx{^b}{M}{_x},  \leftidx{^b}{M}{_z}$} (drone);
\end{tikzpicture}

%% file: figs/control_arch1.tex
\tikzstyle{block} = [rectangle, rounded corners, minimum width=1cm, minimum height=1cm,text centered, draw=black]

\begin{tikzpicture}[node distance = 1.35cm, auto]
\node (pseudo)[]{};
\node (pc) [block, right =of pseudo, align=center] {Position \\Controller};
\node (vc) [block, right =of pc, align=center] {Velocity \\Controller};
\node (ac) [block, right =of vc, align=center] {Acceleration \\ Mixer};
\node (yc) [block, below =of vc, align=center] {Yaw \\ Controller};
\node (ptc) [block, right =of ac, align=center] {Pitch \\ Controller};
\node (drone) [block, below =of ptc, align=center] {Drone};

\draw [->] (pseudo) -- node[anchor=south east]
{$\leftidx{^w}{p}{_x}, \leftidx{^w}{p}{_y}$} (pc);
\draw [->] (pseudo) -- node[anchor=north east]{$\tan^{-1}(^we_y/^we_x)$} (yc);
\draw [->] (yc) -- node[anchor=south east] {$\leftidx{^r}{M}{_z}$} (ac);
\draw [->] (pc) -- node[anchor=south] {$\leftidx{^r}{v}{_x}$} (vc);
\draw [->] (vc) -- node[anchor=south] {$\leftidx{^r}{a}{_x}$} (ac);
\draw [->] (ac) -- node[anchor=south] {$\leftidx{^b}{\theta}{_{pitch}}$} (ptc);
\draw [->] (ptc) -- node[anchor=east] {$\leftidx{^b}{M}{_y}$} (drone);
\draw [->] (ac) -- node[anchor=north east] {$|F|, \leftidx{^b}{M}{_x},  \leftidx{^b}{M}{_z}$} (drone);
\end{tikzpicture}